\documentclass{article}
\usepackage[utf8]{inputenc}
\usepackage[T1]{fontenc}
\usepackage{amsmath, amssymb, amsthm}
\usepackage{geometry}
\usepackage{graphicx}
\usepackage{algorithm}
\usepackage{algorithmic}
\usepackage{booktabs}
\usepackage{hyperref}
\usepackage{xcolor}
\usepackage{listings}
\usepackage{natbib}
\definecolor{codegray}{gray}{0.95}
\definecolor{commentgreen}{rgb}{0.13, 0.54, 0.13}
\definecolor{keywordblue}{rgb}{0.13, 0.13, 1}

{}
{}
{}
{}

\newcommand{\FB}{\text{GFT}_{\text{FB}}}
\newcommand{\RO}{\text{GFT}_{\text{RO}}}
\newcommand{\SO}{\text{GFT}_{\text{SO}}}
\newcommand{\BO}{\text{GFT}_{\text{BO}}}

\definecolor{codegreen}{rgb}{0,0.6,0}
\definecolor{codegray}{rgb}{0.5,0.5,0.5}
\definecolor{codepurple}{rgb}{0.58,0,0.82}
\definecolor{codeblue}{rgb}{0.0,0.0,0.6}
\title{\bfseries A New Lower Bound for the Random Offerer Mechanism in Bilateral Trade using AI-Guided Evolutionary Search} 

\author{
Yang Cai\thanks{Authors are alphabetically ordered. Part of this work was done while Yang Cai was visiting Google.}\\
Yale University\\
\tt{yang.cai@yale.edu}
\and 
Vineet Gupta\\
Google DeepMind\\
\tt{vineet@google.com}
\and
Zun Li\\
Google DeepMind\\
\tt{lizun@google.com}
\and
Aranyak Mehta\\
Google Research\\
\tt{aranyak@google.com}
}
\date{}

\begin{document}

\maketitle

\begin{abstract}
The celebrated Myerson--Satterthwaite theorem shows that in bilateral trade, no mechanism can be simultaneously fully efficient, Bayesian incentive compatible (BIC), and budget balanced (BB). This naturally raises the question of how closely the gains from trade (GFT) achievable by a BIC and BB mechanism can approximate the first-best (fully efficient) benchmark. The optimal BIC and BB mechanism is typically complex and highly distribution-dependent, making it difficult to characterize directly. Consequently, much of the literature analyzes simpler mechanisms such as the Random-Offerer (RO) mechanism and establishes constant-factor guarantees relative to the first-best GFT. An important open question concerns the worst-case performance of the RO mechanism relative to first-best (FB) efficiency. While it was originally hypothesized that the approximation ratio $\frac{\FB}{\RO}$ is bounded by $2$, recent work provided counterexamples to this conjecture:~\cite{cai2021multi} proved that the ratio can be strictly larger than $2$, and~\cite{babaioff2021note} exhibited an explicit example with ratio approximately $2.02$.

In this work, we employ AlphaEvolve, an AI-guided evolutionary search framework, to explore the space of value distributions. We identify a new worst-case instance that yields an improved lower bound of $\frac{\FB}{\RO} \ge \textbf{2.0749}$. This establishes a new lower bound on the worst-case performance of the Random-Offerer mechanism, demonstrating a wider efficiency gap than previously known.

\end{abstract}

\section{Introduction}
The fundamental limits of efficiency in bilateral trade are a central question in mechanism design. In the standard bilateral-trade setting with independent private values, the environment consists of a single seller and a single buyer: the seller has a private cost $s$ for parting with the item, drawn from a distribution $F_s$, and the buyer has a private value $b$ for acquiring the item, drawn independently from a distribution $F_b$. The goal is to design a mechanism that maximizes efficiency, measured by gains from trade (GFT): trade yields $b-s$ if it occurs and $0$ otherwise (equivalently, $\max\{b-s,0\}$ under the efficient allocation).
The Myerson--Satterthwaite theorem establishes that no mechanism can be simultaneously fully efficient, Bayesian incentive compatible (BIC), individually rational (IR), and budget balanced (BB). Consequently, a long line of work studies how well BIC, IR, and BB mechanisms can approximate the first-best (fully efficient) GFT.

The optimal BIC, IR, and BB mechanism is typically complex and highly distribution-dependent, making it difficult to characterize directly. Consequently, much of the literature analyzes simpler mechanisms such as the Random-Offerer (RO) mechanism and establishes constant-factor guarantees relative to the first-best GFT. An important open question concerns the worst-case performance of the RO mechanism relative to the first-best (FB) benchmark. The metric of interest is the approximation ratio
\begin{equation}
    \rho \;=\; \frac{\FB}{\RO}\,.
\end{equation}
The goal is to determine the worst-case value of $\rho$, i.e., the supremum of $\rho$ over all distribution pairs $(F_s, F_b)$.

While it was originally hypothesized that this approximation ratio is bounded by 2, recent work provided counterexamples to this conjecture:~\cite{cai2021multi} proved that the ratio can be strictly larger than $2$, and~\cite{babaioff2021note} exhibited an explicit example with ratio approximately $2.02$. 

In this work, we employ AlphaEvolve, an AI-guided evolutionary search framework, to explore the space of valuation distributions. By reformulating the search for worst-case distributions as a program synthesis problem, we utilize a Large Language Model (LLM) coding agent to evolve distribution structures that maximize the efficiency gap.

We report the discovery of a new worst-case seller distribution. We identify a \emph{mixture of modulated power laws} with cumulative distribution function
\begin{equation}
    \Pr[s \le m]
    \;=\;
    0.2\, z_m^{\alpha_{\mathrm{eff}}(z_m)}
    \;+\;
    0.8\, z_m^{4},
\end{equation}
where $z_m := \frac{m+1}{H+1}$ and the (sinusoidally) modulated exponent is
\begin{equation}
    \alpha_{\mathrm{eff}}(z)
:=    0.15 + 0.05 \sin(2\pi z),
\end{equation}
with $H=20000$.

After discretizing the seller distribution (by rounding the induced probability mass function) 
and pairing it with the Discrete Equal Revenue distribution for the buyer, 
the resulting instance achieves an approximation ratio of $\mathbf{2.0749}$ 
against the Random Offerer mechanism. 
Consequently, $2.0749$ is a new lower bound on the mechanism’s worst-case approximation ratio.

As an illustration of the asymmetry between the sub-mechanisms, our worst-case instance yields 
an expected First-Best GFT of $\FB \approx 1.2322$. 
The individual sub-mechanisms yield a Seller-Offering GFT of $\SO \approx 0.3312$ 
and a Buyer-Offering GFT of $\BO \approx 0.8565$. 
This results in a combined Random Offerer GFT of $\RO \approx 0.5939$. 
Consequently, the approximation ratio is $\rho \approx 2.0749$, 
while the ratio between the First-Best GFT and the maximum of the two sub-mechanisms 
is approximately $1.4387$. 
In contrast, the previous state-of-the-art only shows that the ratio between the First-Best GFT 
and $\max\{\BO,\SO\}$ is at most $4/3$~\citep{babaioff2021note}.

\section{Related Work}

\paragraph{Efficiency in Bilateral Trade.}

Motivated by the impossibility result of \citet{MS83}, there is a growing line of research on designing mechanisms that provably approximate the optimal gains-from-trade (GFT)~\citep[e.g.,][]{blumrosen2016approximating,BCWZ17,DengMSW22,fei2022improved,hartline2025geometric,deng2025approximately} in the bilateral trade setting, as well as in more general settings such as double auctions and multi-dimensional two-sided markets~\citep[e.g.,][]{BabaioffCGZ18,cai2021multi}.
A complementary line of work designs mechanisms that approximate the optimal welfare, in either bilateral trade~\citep[e.g.,][]{BlumrosenD21,KangPV22,CaiW23,dobzinski2024bilateral,dobzinski2025bilateral} or more general two-sided markets~\citep[e.g.,][]{Colini-Baldeschi16,colini2020approximately,DuttingRT14,dutting2026efficient}.
Although exactly optimizing welfare is equivalent to optimizing GFT, approximating welfare is strictly easier than approximating GFT. Indeed, any $c$-approximation to GFT is automatically a $c$-approximation to welfare, whereas the opposite direction is generally false. In this paper, we focus on the approximability of GFT.

\paragraph{Performance of the Random Offerer Mechanism.}
The Random Offerer (RO) mechanism was introduced by~\citet{BCWZ17}, who showed that RO achieves at least half of the second-best GFT. Since the second-best GFT is often a complicated object, researchers frequently use RO as a simple proxy.

\begin{itemize}
    \item The breakthrough of~\citet{DengMSW22}, which shows that the first-best GFT is at most a constant factor larger than the second-best GFT in bilateral trade, builds on RO. In particular, they proved that RO guarantees at least a $1/8.23$ fraction (approximately $0.121$) of the first-best gains from trade.
    \item \citet{fei2022improved} subsequently tightened this bound, improving the guarantee to $1/3.15$.
\end{itemize}

\paragraph{AI guided algorithm design using AlphaEvolve}

    The AlphaEvolve system is presented in~\cite{novikov2025alphaevolve}. 
AlphaEvolve is a significant enhancement of the FunSearch system described in~\cite{RomeraParedes2024FunSearch}. 
\cite{novikov2025alphaevolve} describe new algorithms discovered by AlphaEvolve across diverse domains, including faster matrix multiplication, various mathematical problems, and datacenter optimization. 
Subsequent work has leveraged AlphaEvolve to address problems in mathematics and theoretical computer science. 
For example,~\citep{georgiev2025mathematical} matched the best known solutions for a large number of mathematical problems and improved several of them. 
\citep{nagda} used AlphaEvolve to discover combinatorial structures that yield new results in complexity theory for the Max-Cut and Traveling Salesman problems.

\section{Problem Formulation}

\subsection{Bilateral Trade Setting}
We consider a standard bilateral trade setting with independent private values. The environment consists of a single seller with a cost $s$ and a single buyer with a valuation $b$.
\begin{itemize}
    \item The seller's cost $s$ is drawn from a probability distribution $F_s$ with support in the domain $\Omega$.
    \item The buyer's valuation $b$ is drawn from a probability distribution $F_b$ with support in the domain $\Omega$.
\end{itemize}
We assume the distributions are independent and known to the mechanism, while the realizations $s$ and $b$ are private information.

\subsection{Efficiency Benchmarks}
The metric of interest is the expected \textit{Gains from Trade} (GFT), defined as the expected surplus generated by the transaction.

\subsubsection{First-Best Efficiency (FB)}
The First-Best benchmark represents the maximum theoretically achievable surplus, where trade occurs if and only if $b \ge s$. The expected First-Best GFT is given by:
\begin{equation}
    \text{GFT}_{\text{FB}}(F_s, F_b) = \mathbb{E}_{s \sim F_s, b \sim F_b} \left[ (b - s)^+ \right] = \iint_{b \ge s} (b - s) \, dF_s(s) \, dF_b(b)
\end{equation}

\subsubsection{The Random Offerer Mechanism (RO)}
The Random Offerer mechanism operates by randomizing between two sub-mechanisms with equal probability:
\begin{enumerate}
    \item \textbf{Seller-Offering (SO):} With probability $0.5$, the seller proposes a take-it-or-leave-it price $p_s$. The seller chooses $p_s$ to maximize their expected profit given the buyer's distribution:
    \[
    p_s^*(s) = \arg\max_{p} \left( (p - s) \cdot \Pr_{b\sim F_b}[b\geq p] \right)
    \]
    Trade occurs if $b \ge p_s^*(s)$. The expected GFT for this sub-mechanism is:
    \[
    \text{GFT}_{\text{SO}} = \mathbb{E}_{s, b} \left[ (b - s) \cdot \mathbb{I}(b \ge p_s^*(s)) \right]
    \]

    \item \textbf{Buyer-Offering (BO):} With probability $0.5$, the buyer proposes a take-it-or-leave-it price $p_b$. The buyer chooses $p_b$ to maximize their expected profit given the seller's distribution:
    \[
    p_b^*(b) = \arg\max_{p} \left( (b - p) \cdot \Pr_{s\sim F_s}[s\leq p] \right)
    \]
    Trade occurs if $s \le p_b^*(b)$. The expected GFT for this sub-mechanism is:
    \[
    \text{GFT}_{\text{BO}} = \mathbb{E}_{s, b} \left[ (b - s) \cdot \mathbb{I}(s \le p_b^*(b)) \right]
    \]
\end{enumerate}

The total expected GFT of the Random Offerer mechanism is:
\begin{equation}
    \text{GFT}_{\text{RO}}(F_s, F_b) = \frac{1}{2} \text{GFT}_{\text{SO}} + \frac{1}{2} \text{GFT}_{\text{BO}}
\end{equation}

\subsection{Approximation Ratio and Objective}
We define the approximation ratio $\rho$ as the ratio of the First-Best efficiency to the efficiency of the Random Offerer mechanism. We aim to identify the worst-case distributions that maximize this ratio.

The optimization problem is defined as finding the supremum of $\rho$ over the space of all possible distribution pairs $(F_s, F_b)$:
\begin{equation}
    \rho^* = \sup_{F_s, F_b} \frac{\text{GFT}_{\text{FB}}(F_s, F_b)}{\text{GFT}_{\text{RO}}(F_s, F_b)}
\end{equation}

\section{Methodology: AI-Guided Discovery}

To explore the limits of bilateral trade efficiency, we employ \textbf{AlphaEvolve}, a search framework driven by a Large Language Model (LLM) coding agent. This approach reformulates the search for worst-case distributions as a program synthesis problem. Rather than optimizing numerical parameters within a pre-defined function, the agent evolves Python code to discover novel distribution structures that maximize the efficiency gap.

\subsection{Search Configuration}

Our optimization objective is to find the lower bound of the approximation ratio $\rho$, defined as the ratio of First-Best (FB) efficiency to the Random Offerer (RO) efficiency. To strictly target the worst-case performance, we configure the search space as follows:

\paragraph{Fixed Buyer Distribution}
Following the counterexample structure provided by~\cite{babaioff2021note}, we fix the buyer's valuation to the \textbf{Discrete Equal Revenue Distribution}. The buyer's survival function is held constant throughout the search:
\begin{equation}
    \Pr(b \ge m) = \frac{1}{m} \quad \text{for } m \in \{1, \dots, H\}.\quad  \Pr(b\ge H+1)=0
\end{equation}
where $H$ is the highest possible value.

\paragraph{Evolving Seller Distribution}
The search focuses exclusively on the seller's valuation distribution $F_s$. The evolutionary agent iteratively modifies the source code of a generator function, \texttt{get\_seller\_distributions()}, which constructs the seller's Cumulative Distribution Function (CDF).

\subsection{Evolutionary Process}

The search process operates as an iterative improvement loop:

\begin{enumerate}
    \item \textbf{Initialization:} To avoid biasing the search toward complex pre-conceived structures, we use a uniform distribution as the initial distribution:
    \begin{equation}
        \Pr(s\le m) = \frac{m+1}{H+1}\quad\text{for } m \in \{1, \dots, H\}
    \end{equation}
    \item \textbf{Code Evolution:} In each generation, the LLM agent proposes mutations to the Python code. These modifications can range from simple parameter tuning to the introduction of non-linear functional forms (e.g., power laws, sinusoidal modulation).
    \item \textbf{Fitness Evaluation:} Each candidate program is executed to generate a discrete distribution over the domain $\{0, \dots, H\}$. The fitness of the candidate is strictly defined as the resulting approximation ratio.
\end{enumerate}

\subsection{Evaluation and Numerical Precision}

A critical challenge in identifying worst-case distributions is distinguishing genuine theoretical gains from floating-point errors, particularly when ratios differ by small margins (e.g., $10^{-3}$). To ensure the validity of our results, we implemented a rigorous evaluation pipeline:

\begin{itemize}
    \item \textbf{Discrete Domain:} We utilize a discrete integer domain up to $H=20,000$. This large support allows for fine-grained approximation of continuous structures while maintaining computational tractability.
    
\item \textbf{Rounding and High-Precision Arithmetic:}
To avoid numerical issues, we round the probability mass functions to multiples of a small $\varepsilon=10^{-15}$ and perform the computation using integer arithmetic rather than floating-point operations. 
    
    \item \textbf{Exact GFT Computation:} We compute the Gains from Trade (GFT) exactly.
    \begin{itemize}
        \item For \textbf{First-Best (FB)}, we sum the surplus over all disjoint pairs $(s, b)$ where $b \ge s$.
        \item For \textbf{Random Offerer (RO)}, we explicitly solve for the optimal reserve prices $p^*_s(s)$ and $p^*_b(b)$ for every possible realization of $s$ and $b$, respectively. The final RO efficiency is the exact average of the Seller-Offering and Buyer-Offering sub-mechanisms.
    \end{itemize}
\end{itemize}

This rigorous numerical approach ensures that the reported approximation ratio of \textbf{2.0749} represents a verified lower bound on the mechanism's efficiency. Due to rounding, however, our analysis only establishes this improved lower bound for a rounded version of the mixture of modulated power laws. Determining the exact value of $\rho$ for the unrounded mixture of modulated power laws remains open.

\subsection{Other attempts}
We also experimented with two other evolutionary approaches:
\begin{itemize}
    \item Starting from the solution obtained via the process above, we fix the seller's distribution and evolve the buyer's distribution, initialized with the equal revenue distribution. This approach led to a slight improvement (2.08), which we do not describe in detail here. One could imagine continuing this process by alternating between the two sides whenever progress stalls; we leave such explorations for future work.
    
    \item We also ran experiments evolving both the seller and buyer distributions simultaneously. This produced a new lower bound as well, although it did not surpass 2.07 and progress was significantly slower. The slow progress in this setting ultimately led us to the approach adopted in this paper: fixing the buyer's distribution and evolving only the seller's distribution.
\end{itemize}

\section{Results}

Our application of the AlphaEvolve framework has successfully identified a new counterexample that establishes a tighter lower bound on the worst-case approximation ratio of the Random Offerer mechanism.

\subsection{Main Result}
We report a worst-case approximation ratio of \textbf{2.0749}. This result is achieved with a discretization domain of $H=20,000$.
The specific expected Gains from Trade (GFT) components that yield this ratio are detailed below:
\begin{itemize}
    \item \textbf{First-Best GFT} ($\FB$): 1.2322
    \item \textbf{Seller-Offering GFT} ($\SO$): 0.3312
    \item \textbf{Buyer-Offering GFT} ($\BO$): 0.8565
    \item \textbf{Random Offerer GFT} ($\RO$): 0.5939
    \item \textbf{Approximation Ratio} ($\rho$): 2.0749
\end{itemize}

This finding improves upon the previous best-known lower bound of $\approx 2.02$ established by~\cite{babaioff2021note}. 

\subsection{Discovered Distributions}
The configuration yielding this ratio consists of the fixed buyer distribution (Discrete Equal Revenue) paired with a novel, evolved seller distribution.

\subsubsection{Seller Distribution Structure}
The evolutionary search converged on a \textbf{Mixture of Modulated Power Laws}. Unlike standard power-law distributions used in previous counterexamples, the discovered distribution $F_s$ employs a sinusoidal modulation of the exponent. 

For a normalized domain value $z_m = \frac{m+1}{H+1}$, the Cumulative Distribution Function (CDF) is defined as:
\begin{equation}
    F_s(m) = w \cdot z_m^{\alpha_{\text{eff}}(z_m)} + (1-w) \cdot z_m^{\alpha_2}
\end{equation}
where:
\begin{itemize}
    \item The mixing weight is $w = 0.20$.
    \item The secondary component is a standard power law with $\alpha_2 = 4.0$.
    \item The primary component uses a modulated exponent:
    \begin{equation}
        \alpha_{\text{eff}}(z_m) = 0.15 + 0.05 \sin(2\pi z_m)
    \end{equation}
\end{itemize}

\begin{figure}[t]
    \centering
    \includegraphics[width=0.5\linewidth]{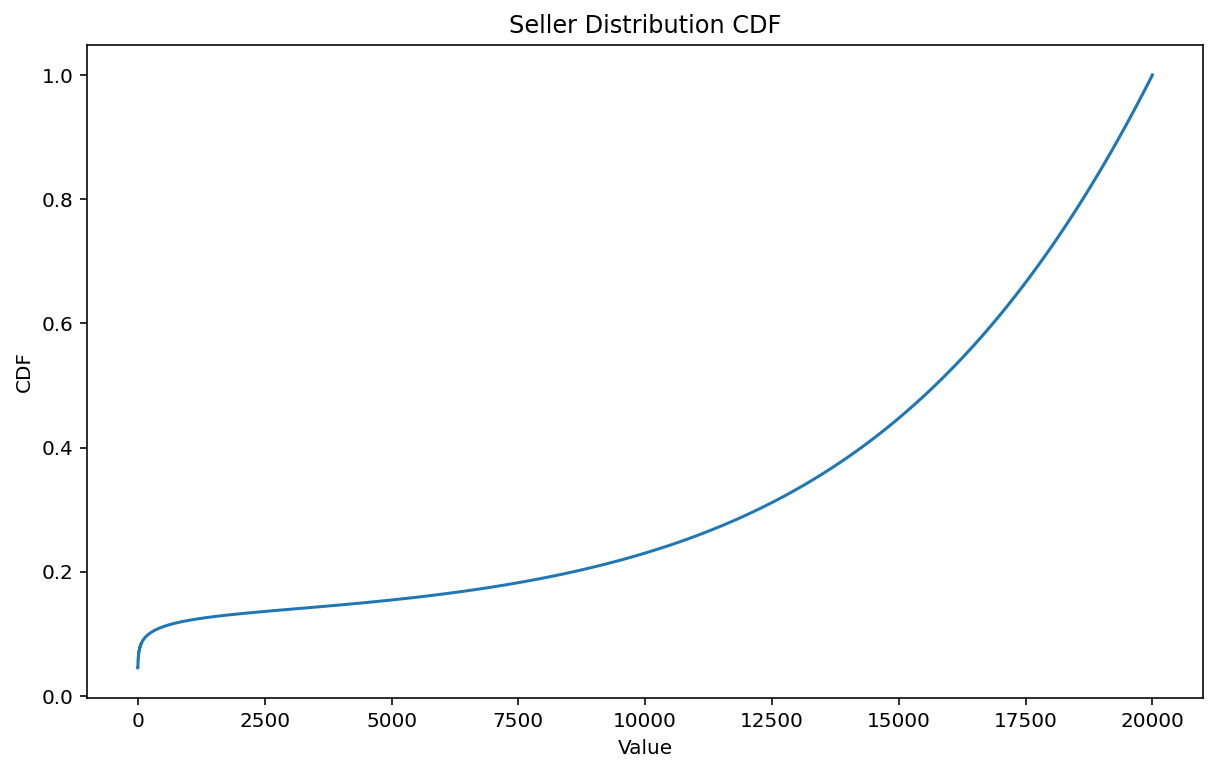}
    \caption{Seller's distribution as found by AlphaEvolve}
    \label{fig:seller-cdf}
\end{figure}

The distribution is shown in Figure~\ref{fig:seller-cdf}. 

\subsection{Evolved Code Artifact}
The Python code generated by the AlphaEvolve agent that constructs this distribution is presented below. The algorithm discovered that introducing the \texttt{math.sin} function to modulate the power law exponent (\texttt{a1}) was critical for maximizing the ratio.

\begin{lstlisting}[caption={Discovered Seller Distribution.}, label={lst:seller}, language=Python]
def get_seller_distributions():
    # Domain size for evaluation
    H = 20000 
    
    # Evolved Parameters
    a1_base = 0.15  # Base exponent for first component
    a2 = 4.0        # Exponent for second component
    w = 0.20        # Weight for first component
    
    # Sinusoidal modulation parameters discovered by AlphaEvolve
    a1_amp = 0.05   # Amplitude of modulation
    a1_freq = 2.0   # Frequency (cycles over domain)

    cdf_s = {}
    norm_factor = H + 1.0
    prev_cdf = 0.0

    for m in range(H + 1):
        # Normalized support z in [0, 1]
        base = max(0.0, min(1.0, (m + 1.0) / norm_factor))

        # Modulate the exponent a1 sinusoidally
        # The argument to sin scales to 2*pi over the domain
        a1_eff = a1_base + a1_amp * math.sin(a1_freq * math.pi * base)
        a1_eff = max(1e-9, a1_eff) # Safety floor

        # Compute mixture components
        cdf1_val = base ** a1_eff
        cdf2_val = base ** a2
        
        # Weighted mixture
        current_cdf = w * cdf1_val + (1.0 - w) * cdf2_val

        # Enforce constraints
        current_cdf = max(0.0, min(current_cdf, 1.0))
        current_cdf = max(current_cdf, prev_cdf) # Monotonicity

        cdf_s[m] = current_cdf
        prev_cdf = current_cdf

    return H, cdf_s
\end{lstlisting}

The discovery of the sinusoidal modulation parameter \texttt{a1\_amp = 0.05} highlights the capability of the evolutionary search to identify non-intuitive functional forms that human theoretical analysis might overlook.
\section{Conclusion}

In this work, we revisited the fundamental efficiency limits of the Random Offerer (RO) mechanism in bilateral trade. By deploying \textbf{AlphaEvolve}, an AI-guided evolutionary search framework, we successfully identified a new lower bound on the worst-case approximation ratio.

Our primary contribution is the discovery of a specific valuation distribution pair that yields an approximation ratio of \textbf{2.0749}. This result widens the best previous efficiency gap of $\approx 2.02$ provided by~\cite{babaioff2021note}. The structure of the discovered seller distribution—a mixture of power laws modulated by a sinusoidal function is significantly different from the previous worst-case distribution, and may suggest new ways that the Random Offerer mechanism can be exploited.

Methodologically, this study highlights the potential of large language models as coding agents in microeconomic theory. 
AlphaEvolve’s ability to synthesize non-intuitive functional forms—such as the sinusoidally modulated exponent identified in our results—suggests that AI-driven search can be a powerful tool for probing the limits of mechanism design. 
This approach may be applicable to other open problems in auction theory and algorithmic game theory, where analytical derivations of worst-case bounds remain elusive.

\bibliographystyle{plainnat}

\bibliography{references}

\appendix
\section{Evaluation Code}
\begin{lstlisting}[caption={Evaluation Code.}, label={lst:eval}]
import numpy as np
try:
  DTYPE = np.float128
  # On some systems, float128 is just an alias for float64. This checks for that.
  if np.finfo(DTYPE).nexp == np.finfo(np.float64).nexp:
    DTYPE = np.float64
except AttributeError:
  DTYPE = np.float64




def _derive_pmf_from_integer_cdf(cdf, H):
  """Helper to derive PMF from CDF using NumPy types."""
  pmf = {m:0 for m in range(H + 1)}
  pmf[0] = cdf.get(0, 0)
  for m in range(1, H + 1):
    # Ensure default value is of the correct type
    pmf[m] = cdf.get(m, 0) - cdf.get(m - 1, 0)
    if pmf[m] < 0:
      pmf[m] = 0
  return pmf


def _derive_pmf_from_integer_sf(sf, H):
  """Helper to derive PMF from Survival Function using NumPy types."""
  pmf = {m: 0 for m in range(H + 1)}
  for m in range(H):
    pmf[m] = sf.get(m, 0) - sf.get(m + 1, 0)
    if pmf[m] < 0:
      pmf[m] = 0
  pmf[H] = sf.get(H, 0)
  return pmf


def _round_to_integer_distribution(cdf_s, sf_b, H, rtol):

  if not rtol > 0:
    raise ValueError("Rounding tolerance 'rtol' must be a positive number.")

  # Use dictionary comprehensions for a concise and efficient implementation.
  # For each item in the input dictionary, the key (valuation) is kept,
  # and the value (probability) is rounded to the nearest multiple of rtol.
  # The type of the result will be preserved as DTYPE.
  rounded_cdf_s = {
      val: round(prob / rtol)
      for val, prob in cdf_s.items()
  }

  rounded_sf_b = {
      val: round(prob / rtol)
      for val, prob in sf_b.items()
  }

  for m in range(H + 1):
    rounded_cdf_s[m] = rounded_cdf_s.get(m, 0)
    rounded_sf_b[m] = rounded_sf_b.get(m, 0)

  return rounded_cdf_s, rounded_sf_b


def calculate_all_gains_from_trade_rounded_integer_fixed_seller(H, sf_b, cdf_s):

    ### Make sure support is not too large
  if H > 20000:
    return DTYPE(-1)

  ### The number of digits to keep after the decimal point
  tolerance = DTYPE("1e-15")

  ### Get fixed seller distribution
  # cdf_s = get_fixed_seller_distribution(H)

  ### Round the CDF for seller and SF for buyer to integer multiples of tolerance, add missing entries, and check whether the CDF/SF's are valid
  cdf_s_np, sf_b_np = _round_to_integer_distribution(cdf_s, sf_b, H, tolerance)


  ### Check if the integer distributions have valid monotone cdf and sf.
  # _validate_integer_distributions(H, cdf_s_np, sf_b_np)

  pmf_s = _derive_pmf_from_integer_cdf(cdf_s_np, H)
  pmf_b = _derive_pmf_from_integer_sf(sf_b_np, H)

  first_best_gft = 0
  for s_val in range(H + 1):
    if pmf_s[s_val] <= 0:
      continue
    for b_val in range(s_val, H + 1):
      if pmf_b[b_val] <= 0:
        continue
      gain = b_val - s_val
      first_best_gft += pmf_s[s_val] * pmf_b[b_val] * gain

  gft_seller_offering = 0
  for s_val in range(H + 1):
    if pmf_s[s_val] <= 0:
      continue
    max_seller_profit = 0
    p_s_opt = s_val
    for p_offer in range(s_val, H + 1):
      prob_buyer_accepts = sf_b_np.get(p_offer, 0)
      current_seller_profit = prob_buyer_accepts * (p_offer - s_val)
      if current_seller_profit >= max_seller_profit:
        max_seller_profit = current_seller_profit
        p_s_opt = p_offer
    expected_gft_for_this_s = 0
    for b_val in range(p_s_opt, H + 1):
      if pmf_b[b_val] <= 0:
        continue
      gain = b_val - s_val
      expected_gft_for_this_s += pmf_b[b_val] * gain
    gft_seller_offering += pmf_s[s_val] * expected_gft_for_this_s

  gft_buyer_offering = 0
  for b_val in range(H + 1):
    if pmf_b[b_val] <= 0:
      continue
    max_buyer_profit = 0
    p_b_opt = b_val
    for p_offer in range(b_val + 1):
      prob_seller_accepts = cdf_s_np.get(p_offer, 0)
      current_buyer_profit = prob_seller_accepts * (b_val - p_offer)
      if current_buyer_profit > max_buyer_profit:
        max_buyer_profit = current_buyer_profit
        p_b_opt = p_offer
    expected_gft_for_this_b = 0
    for s_val in range(p_b_opt + 1):
      if pmf_s[s_val] <= 0:
        continue
      gain = b_val - s_val
      expected_gft_for_this_b += pmf_s[s_val] * gain
    gft_buyer_offering += pmf_b[b_val] * expected_gft_for_this_b

  gft_random_offerer = DTYPE(0.5) * (gft_seller_offering + gft_buyer_offering)
  return first_best_gft/gft_random_offerer



\end{lstlisting}
\end{document}